\pdfoutput=1

\documentclass[11pt]{article}

\usepackage{acl}

\usepackage{times}
\usepackage{latexsym}

\usepackage[T1]{fontenc}

\usepackage[utf8]{inputenc}

\usepackage{microtype}

\usepackage{inconsolata}
\usepackage{booktabs}
\usepackage{amsmath}
\usepackage[inline]{enumitem}

\usepackage{graphicx}
\graphicspath{ {./images/} }

\title{Robustness of Large Language Models to Perturbations in Text}

\author{Ayush Singh, Navpreet Singh, Shubham Vatsal  \\
  inQbator AI at eviCore Healthcare \\
  Evernorth Health Services \\
  \texttt{firstname.lastname@evicore.com}}
  
\begin{document}
\maketitle
\begin{abstract}
Having a clean dataset has been the foundational assumption of most natural language processing (NLP) systems. However, properly written text is rarely found in real-world scenarios and hence, oftentimes invalidates this foundational assumption. Recently, large language models (LLMs) have achieved remarkable performance in a wide array of NLP tasks. Nevertheless, the degree to which these LLMs are robust to semantic preservation of morphological variations in text has been sparsely studied. In a world becoming increasingly dependent on LLMs for most of the NLP tasks, it becomes crucial to know their robustness to numerous forms of noise found in real-world text. In this work, we systematically evaluate LLMs' resilience to corrupt variations of the original text. We do so by artificially introducing different levels of noise into the discussed datasets. We show that contrary to popular beliefs, generative LLMs are quiet robust to commonly found perturbations in text. 
Additionally, we test LLMs' performance on multiple benchmarks achieving a new state of the art on the task of grammar error correction. To empower future research, we are also releasing a dataset annotated by humans stating their preference for LLM vs. human-corrected outputs along with the code to reproduce our results.
\end{abstract}
\section{Introduction}
Modern day language processing (NLP) pipelines have been highly dependent on the input data to be as clean as possible. While this assumption works well in well-curated settings, it often gets invalidated in real-world setting leading to brittle systems that work well in laboratories, albeit, breaking down on the noisy naturally occurring data \cite{wu2021bridging}.
Consequently, there emerges a need to test any new innovation in the field of NLP on it's robustness against several different forms of noise that are pervasive in real-world scenarios \cite{galliers1993evaluating}. 

Noise in real-world datasets can originate from a plethora of resources. Some of the errors can originate from human factors like spelling or grammatical errors while others can be machine induced like errors from optical character recognition (OCR), or automated speech recognition (ASR) systems. These various forms of noise have been a significant impediment deploying systems built on cleaner datasets in a real word setting. For example, even notes taken by native speakers oftentimes include spelling and grammatical mistakes, while text written by non-native speakers miss determiners and exhibit numerous forms of valid but orthogonal variations. Their impact on downstream performance can vary in degree ranging from a slight change in prediction probability to completely flipping the polarity or semantic meaning of a text (See Table~\ref{tab:examples} for examples). The field of NLP that deals with detecting when a meaning has changed or shifted is known as Lexical Semantic Change (LSC) detection \cite{gulordava2011distributional}. Although LSC approaches help detect this shift, they do not offer the types of change that can be made to shift from one semantic to another such as translating an incorrect text into a lexical and grammatically correct one without changing the meaning.

\begin{table*}[ht]
\centering
\begin{tabular}{lp{2.5in}p{2.5in}}
\textbf{Type} & \textbf{Error} & \textbf{Correct}\\ \midrule
Agreement     & He enjoys reading novels, play chess, and watch movies on weekends. & He enjoys reading novels, playing chess, and watching movies on weekends. \\
Determiner    & They enjoy the sushi for dinner & They enjoy sushi for dinner \\
Morphological & I do not swim as well as he do. & I do not swim as well as he does. \\
Multiple      & I sea the see from the seasoar & I saw the sea from the seesaw \\ 
Preposition   & She put the book in the table. & She put the book on the table. \\
Punctuation   & My favorite fruits are apples bananas and oranges & My favorite fruits are apples, bananas, and oranges. \\ 
Syntax        & She the store went to. & She went to the store. \\
Tense/Aspect  & I has been to London last year. & I went to London last year. \\
Unidiomatic   & She was head over heels in the clouds when she received the promotion. & She was on cloud nine when she received the promotion. \\ 
\end{tabular}
\caption{Examples of various types of errors commonly found in real-world natural language.}
\label{tab:examples}
\end{table*}

Traditionally, machine learning (ML) systems have handled noise in text by using data cleaning pipelines comprising multiple phases, the most important one being known as grammar error correction (GEC) phase. \citet{bryant2023grammatical} pointed out that GEC is a misnomer and has lately been more generally referred to as language error correction (LEC) that involves not only grammatical mistakes such as improper subject-verb agreement, but also spelling or type errors and other forms of errors as well. However, LEC is not an easy task, which is why ML has been employed for it. Although ML has progressed the state of LEC, it has not solved it entirely. In the last decade, emergence of advanced methods like subword embeddings has increased robustness to noise and hints at a future where downstream systems are so robust to noise that LEC might not even be needed anymore. Lately, transformer-based large language models (LLMs) have shown great promise in this area. With LLMs being used predominantly in all aspects of NLP, it is imperative to evaluate the robustness of LLMs to the fundamental task of LEC especially when recent work has shown a high degree of sensitivity by these systems to even word-level perturbations \cite{srivastava2020noisy, wang2023large}.

LLMs have shown remarkable performance in most avenues of NLP tasks. This unprecedented gain comes from learning from a large corpus of natural language in an auto-regressive predictive manner. LLMs are trained on a mixture of clean and noisy text which allows them to incorporate robustness towards minor irregularities in text. However, a comprehensive evaluation of the degree to which they are robust to this noise has not been done yet. Tangentially, there has been a resurgence in the application of LLMs to LEC tasks as well \cite{bryant2023grammatical}.

In this work, we systematically measure the degree to which robustness to semantic-preserving corruption holds for LLMs. We define semantic-preserving by one or more set of corruptions that can happen till the point where a human is able to equate original and corrupted text. We measure robustness by differentiating internal LLM encoding of the clean text with that of it's corresponding corrupt version. We corrupt the text synthetically with increasing degree of severity in the form of individual perturbations as well as combinations of them. Controlling the level of corruption allows us to have a better grasp on what aspects of noise affects LLMs. Additionally, we also measure LLMs performance on downstream tasks of LSC and LEC, which captures nuances of commonly found errors in-the-wild. Our contributions are threefold 1) using real and synthetic datasets, we show the extent to which this robustness holds in LLMs against various types of errors 2) report performance of LLMs on downstream LSC and LEC tasks 3) we share a human annotated data on preference of LLM corrected text vs. that of humans themselves.

\section{Related Work}
\paragraph{Noise} in natural language text has been a well-studied area of research for some time now. Research in this domain started with crude categorization of different types and progressed to more recent fine-grained specifications \cite{lopresti2008optical, dey2009studying, passonneau2009reducing, xing2013checker, al2021towards}. \citet{lopresti2008optical} studied the negative effects of OCR system errors on NLP systems while \citet{dey2009studying} studied the negative effects of noise on text mining applications. With the advent of LLMs, \citet{srivastava2020noisy, wang2023large, naplava2021understanding} studied the impact of noisy text on LLMs and showed that it has negative results. However, they did not evaluate the more recent modern-day generative LLMs. 

\paragraph{Lexical Semantic Change.} LSC detection techniques are split into following three categories \begin{enumerate*}[label=(\roman*)]
    \item  semantic vector spaces,
    \item topic distributions, and
    \item sense clusters
\end{enumerate*}. Because LLMs encode information about the meaning of an input into its internal layers called dense representations or popularly known as embeddings, the first category of LSC detection via semantic vector spaces has gained traction recently \cite{gulordava2011distributional, kim2014temporal, xu2015computational, eger2017linearity, hamilton2016cultural, hellrich2016bad}. 
\citet{schlechtweg2020semeval} proposed the 2020 SemEval task on unsupervised lexical semantic change (LSC) detection which spurred a lot of interest in the field. Since then, several methods have been proposed to detect LSC ranging from using distance metrics \cite{rosenfeld2018deep, qiu2022histbert} to differences in contextual dispersion between the two vectors \cite{kisselew2016predicting}. Even though the choice of distance metric depends on the underlying task at hand, the most common metric has been cosine distance. \citet{martinc2019leveraging, montariol-etal-2021-scalable} proved the efficacy of leveraging contextualized embeddings in detecting diachronic semantic shifts on various corpus and different languages. We apply their techniques with more recent generative LLMs. Furthermore, 
\citet{schlechtweg2019wind, shoemark2020room} corroborated the efficacy of this approach by doing a systematic comparison of semantic change detection approaches with embeddings using cosine similarity. For an in-depth review of all approaches, please refer to recent survey by \citet{tahmasebi2021survey}.

\paragraph{Datasets.} In order to measure the performance of the proposed LEC techniques, several benchmark datasets have been created. \citet{yannakoudakis-etal-2011-new} created the first dataset by manually annotating scripts from ESOL examinations into 88 categories. Later, \citet{ng-etal-2014-conll} introduced a CoNLL-2014 Shared Task on Grammatical Error Correction, however, both of these datasets were low in volume and had some inherent problems in the annotation as revealed by \citet{bryant-etal-2019-bea}, who then also introduced a new dataset as well as LEC task named Building Educational Applications-2019. \citet{napoles-etal-2017-jfleg} also introduced an LEC dataset which was much cleaner and operated on sentence level. While datasets created for LEC specifically measure the ability of a system to successfully do LEC, they do not allow the flexibility to measure the correlation of mistakes with that of downstream performance. To that end, synthetically created datasets fill this crucial gap created by organically derived datasets. \citet{ko2023robustness} showed the benefits of more synthetic datasets in the field of NLP so that we can measure certain aspects of models that otherwise would go undetected in organic datasets.  

\paragraph{Large language models for LEC.} Recently, LLM's ability to handle large number of NLP tasks has hinted at its potential success in GEC \cite{raheja2023coedit}. At the same time, there is a large body of research showing LLM's sensitivity to noise but not in a systematic manner \cite{srivastava2020noisy, wang2023large}. Even LLM based LSC models only allow detection of a semantic change whereas there are times when one would also want to also correct the change in one way or another. This is where LEC shines which involves automatically correcting the syntactical errors in a given text as best as possible. Several works have explored using automated methods for LEC, from rule-based heuristics \cite{sidorov2013rule, xing2013checker}, to machine learning models \cite{garg2021towards}, and more recently neural network-based methods \cite{malykh-2019-robust, zhang2022syngec, raheja2023coedit}.  Recently, \citet{bout2023efficient} posed GEC as a sequence-to-sequence task where the encoder takes inappropriate text as input and the decoder decodes the edits that needs to be made. \citet{fang2023chatgpt} evaluated Chat-GPT on LEC and found that it has excellent capabilities in not only English but also multilingual corrections; however, they did not compare against the recent version of GPT, nor did they evaluate against open source models. 

\paragraph{Prompting LLMs.} The primary means of interaction with LLMs have been to prompt it with an instruction. Additionally, \citet{brown2020language} found that supplying examples with the instruction helps improve the performance. This is known as few-shot prompting while the former is called zero-shot as no examples are provided \citet{kojima2022large}. Recently, \citet{wei2022chain} found out that further performance boost can be achieved by asking LLMs to explain how it arrived at an answer, naming it as Chain-of-Thought (CoT) reasoning. Though several research came after the aforementioned three prompting techniques \cite{zhou2022least, zhou2022large}, none of them brought a dramatic improvement.


\section{Methods}
In this section, we elaborate on the LSC detection methods used to assess the degree of the effect of corruption on the LLMs as well as the techniques used to configure LLMs to perform LEC.

\subsection{Problem statement}
We hypothesize that any change in how an LLM encodes a sequence can be measured by calculating the difference in its dense representations. Therefore, we measure the difference between an originally corrupt or noisy text $\vec{x}$ with its corrected counterpart $\vec{y}$.

Our null hypothesis $H_0$, therefore is that the internal encoding of LLM should be different for a text and it's semantically similar but incorrect version. On the other hand, our alternative hypothesis $H_a$ states that there should be no difference in how LLM encodes a text and it's grammatically incorrect version as the LLM has learned to extract semantic meaning despite of problems in how a text is written. To prove our hypothesis, we measure how distant is the embedding of the corrupted text from it's correct counterpart. If this distance is not significant enough then our null hypothesis is rejected.

\subsection{Lexical Semantic Change (LSC) Detection}
In order to measure the similarity of dense representations of $\vec{x}$ and $\vec{y}$, or when a LSC is detected, we use a standard metric called \textit{cosine similarity}. The cosine similarity measures the angular distance of two vectors by first performing their dot product, divided by the product of their lengths as depicted by the following equation:
\begin{equation}
    \begin{aligned}
    Sim_{cosine}(\vec{x}, \vec{y}) &= \frac{\vec{x} \cdot \vec{y}}{|\vec{x}||\vec{y}|} \\
                                   &= \frac{\sum_{i=1}^N x_i \cdot y_i}{\sqrt{\sum_{i=1}^N x_{i}^2}\cdot \sqrt{\sum_{i=1}^N y_{i}^2}}
    \end{aligned}
\end{equation}



\subsection{Perturbations}
\label{sec:perturb}
Even though benchmark LEC datasets have a incorrect version and it's corresponding human corrected version, the benchmarks only capture a certain aspect of error occurring in text. Fortunately, prior research has done the immense job of categorizing the type of errors pervasive in real-world data because of which we can simulate the errors and measure robustness of LLM against it. To that end, among the numerous form of errors, we used domain knowledge and heuristics to rank the errors and selected for most relevant errors. These errors are elaborated as follows:

\paragraph{OCR error} OCR augmenter emulates the common recognition errors, such as substituting the numeral "0" with the characters "o" or "O". This is achieved through a predefined mapping table, which targets the replacement of such recognized characters.

\paragraph{Spelling and keyboard mistakes} Spelling mistake errors are introduced by substituting words with commonly misspelled alternatives, which are stored in a predefined mapping table. Keyboard input errors emulates mistyping by replacing random characters with those located within one keyword distance away on the QWERTY keyboard layout.

\paragraph{Split, swap and delete word} Split augmenter randomly splits the word into two separate sub words. Swap augmenter randomly exchanges adjacent words within the text. The delete augmenter removes words randomly simulating the occurrence of missing or omitted content.

\paragraph{Contextual insert and substitute word} The contextual augmenter leverages prominent word embedding models like Word2Vec \cite{mikolov2013distributed} or employing LLM like BERT \cite{devlin2018bert} to identify and substitute or insert similar alternative words, enriching the vocabulary diversity within the text.

\paragraph{Substitute, insert, swap and delete character} This versatile character augmenter performs substitution, insertion, deleting and swapping of random characters throughout the textual input.

\paragraph{Synonym and antonym swap} This augmenter utilizes WordNet and PPDB for the strategic replacement of words with their synonyms or antonyms. It conducts preliminary checks before swapping to ensure the appropriateness of the replacement. Words that serve as determiners (e.g., a, an, the, etc) and words without a synonym or antonym are excluded.

\subsection{Prompting}
\label{sec:prompt}
We evaluate LLMs for LEC task by prompting them to correct the errors in the dataset. Even though there are numerous advanced forms of prompting, however, they have been found to be task and dataset dependent. Therefore, in this work, we kept prompting variations to minimal.
We ask the model to correct the language of the text with as little external knowledge as possible in the following prompt format:





\begin{quote}
\small
You are an English language expert who is responsible for grammatical, lexical and orthographic error corrections given an input sentence. Your job is to fix grammatical mistakes, awkward phrases, spelling errors, etc. following standard written usage conventions, but your corrections must be conservative. Please keep the original sentence (words, phrases, and structure) as much as possible. The ultimate goal of this task is to make the given sentence sound natural to native speakers of English without making unnecessary changes. Corrections are not required when the sentence is already grammatical and sounds natural. 

Here is the input sentence containing errors that needs to be corrected.

Input Sentence:

\#\#\#
\textit{\{input\_sentence\}}
\#\#\#

\end{quote}


\section{Experiments}
We split our experiments into increasingly complex phases starting with single perturbations and ramping the combinations up to five perturbations. For perturbations, we sequentially corrupt where each method's probability of an augmentation is 30\% with a maximum of 10 words that can be operated with either of substitution, insertion or deletion. Additionally, we discard any samples whose unigram Jaccard similarity coefficient less than 0.7 after all the corruptions have taken place. The models we chose for comparing dense representations are latest (fourth) version of GPT \cite{openai2023gpt4} with 8k context window (text-embedding-ada-002), the 7 billion version of an open source model LLaMa 3 \cite{touvron2023llama} with 4k context window (decoder head embedding) and a non-generative model BERT \cite{devlin2018bert} with 512 context window (CLS token embedding). We kept all the hyper-parameters to default as the temperature, frequency penalty, and presence penalty to 0 and max tokens to 1000. All of the aforementioned LLMs are based on similar transformer based architectures \cite{vaswani2017attention}. Additionally, we evaluate LLMs on two recent LEC benchmark datasets that earlier works have not evaluated against.

\subsection{Datasets}
For LEC task, we use two benchmark datasets named JFLEG and BEA-19. The JHU FLuency-Extended GUG corpus (JFLEG) extends the GUG (Grammatical/Ungrammatical) corpus by \citet{heilman-etal-2014-predicting} with a layer of annotation via four human annotators \cite{napoles-etal-2017-jfleg}. The key differentiator with JFLEG is that the corrections were made with inclination for fluency rather than minimal edits. This is done on the GUG corpus which is a cross-section of ungrammatical data, containing sentences written by English language learners with different L1s and proficiency levels. BEA-19 \cite{bryant-etal-2019-bea} was introduced as a shared task in the workshop of Building Educational Applications 2019. BEA dataset contains essays on approximately 50 topics written by more than 300 authors from around the world (including native English as well as British and American undergraduates speakers). 

Apart from including both the aforementioned datasets in our LSC task, we also generate synthetic datasets for LSC, we use the IMDB movie review dataset \cite{maas-etal-2011-learning} and sub-sample 1000 reviews from it. The IMDB dataset was crowd-sourced reviews of movies with varying ratings. Even though parallel rating is available for each review, we did not need it for our usecase. For perturbations, we used the work by \cite{ma2019nlpaug} to generate different semantic preserving variations of a given text.

\begin{table}[h]
    \centering
    \begin{tabular}{ccc} \toprule
         \textbf{Dataset} & \textbf{\# sentence pairs} & \textbf{Scorer} \\ \midrule
         BEA-19 train   & 561,410 & -      \\
         BEA-19 dev     & 2,377   & ERRANT \\ \midrule
         BEA-19 test    & 4,477   & ERRANT \\
         JFLEG          & 747     & GLEU   \\
         IMDB           & 11,495  &  -     \\ \bottomrule
    \end{tabular}
    \caption{Corpus statistics of the datasets used in our experiments.}
    \label{tab:corpus-stats}
\end{table}

\subsection{Annotation}
\cite{bryant2023grammatical} posited that most of the LEC evaluation metrics have been calibrated against human preferences as it is a challenging task to evaluate the quality of a correction computation even when ground truth is present. Taking this into account, we also setup an annotation task as a true measure of LLM's ability to perform GEC. We hypothesized that LLMs might behaviorally perform LEC that might be different from the way humans approach LEC.

We sub-sample records from both the JFLEG and BEA dataset to be corrected by both GPT and humans (for humans, we already have the ground truth). We focus more on JFLEG over BEA as the former has four expert human annotations per data point compared to that of one from BEA in the dev set. Additionally, BEA dataset itself was corrected via the same population that wrote the incorrect sentences, which makes the ground truth annotations of relatively lower quality. Our annotators demographically ranged from all over the world and had graduate degrees in STEM fields. We used the Label Studio platform and provided the following instructions to the annotators:

\begin{itemize}[noitemsep]
    \item For a given incorrectly written English text, please select one of two corrections presented.
    \item Select the one that deems most correct to you semantically, grammatically and syntactically.
    \item When both outputs seem correct, pick the one whose words or syntax is as close to the original incorrect sentence.
    \item When both corrections are same, tick the box that they are same.
\end{itemize}

\section{Results}
\begin{figure}[h]
    \includegraphics[scale=0.22]{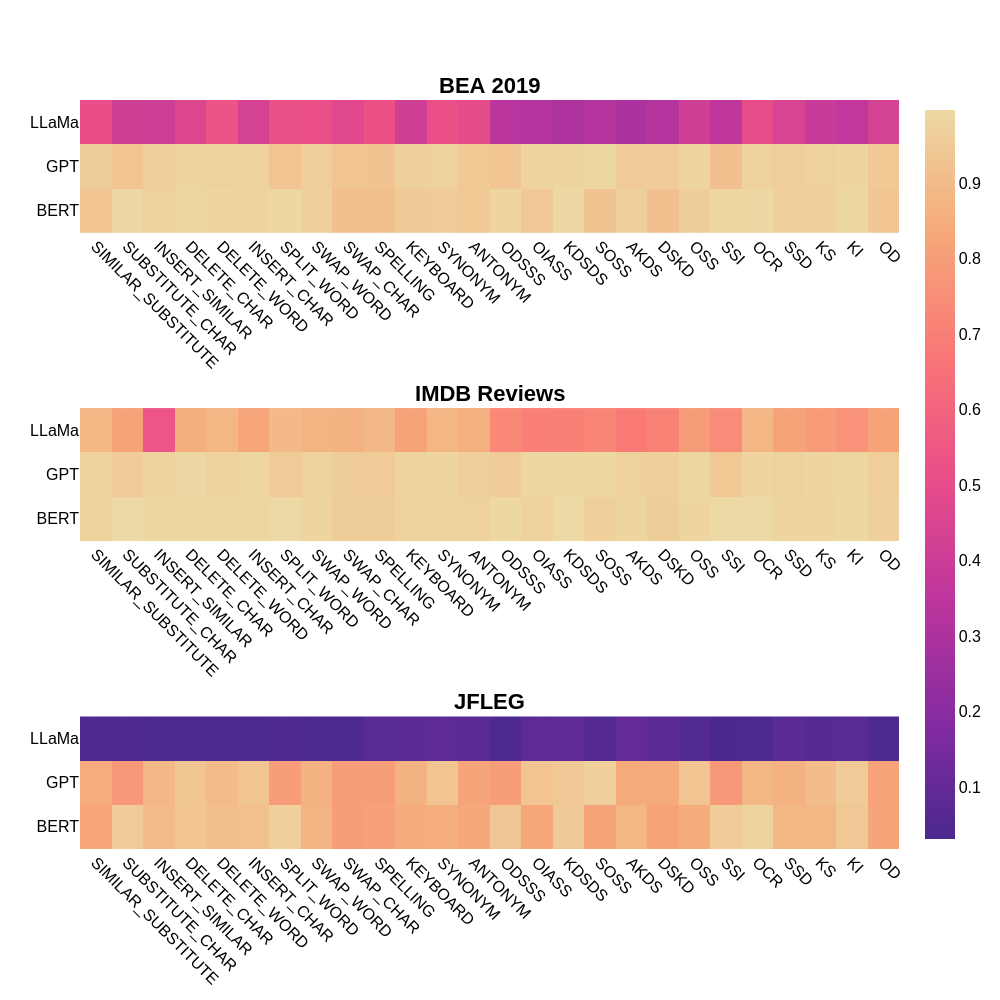}
    \caption{Cosine similarity of clean and corrupted text for all perturbations. Combination's have first character of each perturbation as label (See Section \ref{sec:perturb} for more).}
    \label{fig:heatmap}
\end{figure}

\begin{table}[h]
\centering
\begin{tabular}{@{}lllllll@{}}
\toprule
 & \multicolumn{2}{c}{GPT}  & \multicolumn{2}{c}{LLaMa} & \multicolumn{2}{l}{BERT} \\ 
             \cmidrule(lr){2-3} \cmidrule(lr){4-5} \cmidrule(lr){6-7} 
    Perturbation        & $\mu$                    & $\sigma$ & $\mu$                     & $\sigma$ & $\mu$ & $\sigma$                    \\ \midrule
BEA+1P       & 98  & 1.1 & 50  & 10.6 & 98 & 1.1 \\
BEA+2P       & 95  & 1.3  & 39 & 3.3  & 95 & 1.5 \\
BEA+3P       & 93  & 1.3  & 40 & 4.7  & 95 & 0.4 \\
BEA+4P       & 92  & 0.4  & 31 & 1.6  & 93 & 1.3 \\
BEA+5P       & 91 &  0.9  & 32 & 1.7  & 91 & 0.6 \\ \midrule
IMDB+1P      & 98  & 0.6  & 83 & 9.0  & 99 & 0.5 \\
IMDB+2P      & 97  & 0.7  & 79 & 2.7  & 97 & 0.8 \\
IMDB+3P      & 96  & 1.2  & 78 & 3.9  & 98 & 0.1 \\
IMDB+4P      & 95  & 1.3  & 70 & 1.6  & 96 & 0.8 \\
IMDB+5P      & 94  & 3.4  & 70 & 1.5  & 95 & 0.4 \\ \midrule
JFLEG+1P     & 92  & 3.5  & 11 & 22.9 & 93 & 3.3 \\
JFLEG+2P     & 83  & 1.6  & 6  & 1.4  & 84 & 2.6 \\
JFLEG+3P     & 84  & 2.1  & 5  & 1.9  & 85 & 0.4 \\
JFLEG+4P     & 79  & 0.8  & 8  & 2.4  & 82 & 0.8 \\
JFLEG+5P     & 79  & 0.9  & 6  & 2.7  & 80 & 1.1 \\ \bottomrule
\end{tabular}
    \caption{Mean ($\mu$) and standard deviation ($\sigma$) of cosine similarity of LSC detection task for all three models.}
    \label{tab:results}
\end{table}

As shown in Table \ref{tab:results}, even after severe degradation of the original text, the embeddings of clean and corrupted version of text remain fairly same throughout. This leads us to reject our null hypothesis $H_0$ and accept alternative hypothesis $H_a$ that LLMs are robust to semantic-preserving variations of text. To our surprise, LLaMa did not fair well to different forms of perturbations and the null hypothesis cannot be rejected for LLaMa. Additionally, as can be seen in figure \ref{fig:heatmap}, both BERT \cite{devlin2018bert} and GPT \cite{openai2023gpt4} remain fairly robust to different forms of perturbations whereas LLaMa \cite{touvron2023llama} in some cases simply collapsed on detecting similarity. We elaborate on the worse performance on LLaMa in the discussion section \ref{sec:discuss}.

\begin{table}
    \centering
    \begin{tabular}{lccc} \toprule
        \textbf{Dataset}                       & JFLEG              & BEA-19 test \\ \midrule
        TagGEC \shortcite{stahlberg2021synthetic}   &  64.7         &  70.4             \\
        T5-XXL \shortcite{rothe2021simple}          &    -          &  75.9         \\
        GECToR \shortcite{tarnavskyi2022ensembling} &    58.6          & 73.2          \\
        BART \shortcite{bout2023efficient}          &    -          & \textbf{75.9}  \\
        ChatGPT \shortcite{fang2023chatgpt}         & 61.4          & 36.1           \\ \midrule
        LLaMa3-7b zero-shot                    &      51.9       &      35.2         \\ 
        GPT-4 zero-shot                        & \textbf{64.9} & \textbf{59.6}          \\ \bottomrule
    \end{tabular}
    \caption{Results for both datasets on pre-existing supervised methods and more recent unsupervised methods, where GLEU score is used for JFLEG and ERRANT for BEA-19 test set.}
    \label{tab:lec-results}
\end{table}

Apart from passing the LSC detection test, as shown in Table \ref{tab:lec-results}, the unsupervised GPT approach achieves a new state of art on the JFLEG dataset. On the BEA-19 dataset, GPT surpasses the previous state of art in unsupervised domain i.e. ChatGPT by a significant margin of 23 points or 70\% better. Even LLaMa performed equivalent to ChatGPT being only 1 point behind in $F_{0.5}$ ERRANT score. Nevertheless, unsupervised model's performance still lags behind that of models trained specifically on the BEA-19 dataset. The drastically varying results on both the datasets opens doors to error analysis on why performance soared in one while lagged in the other. We setup some probing and annotation tasks to resolve this in the following section.


\section{Discussion}
\label{sec:discuss}
\begin{table*}[ht]
    \centering
    \small
    \begin{tabular}{ll}
        \textbf{System} & \textbf{Text} \\ \midrule
         Original & New and new technology has been introduced to the society.\\
         Human 1  & New technology has been introduced into the society.\\
         Human 2  & Newer and newer technology has been introduced into society. \\
         GPT     & New technology has been introduced to society. \\
         GPT Formal & Society has witnessed the successive introduction of pioneering technologies. \\ 
         GPT Emotional & The influx of groundbreaking technology into society has been both exhilarating and transformative. \\
         GPT Professional & The continuous integration of new technology into society has been a significant and impactful progression. \\
         GPT Casual & Loads of new tech keep popping up and changing how society rolls. \\\midrule
         
         Original & i have studed for just examination wayse studed. \\
         Human 1  & I have studied for just examination.\\
         Human 2  & I have studied for just examination we studied.\\
         GPT      & I have studied just for the sake of examinations. \\
         GPT Formal & I diligently prepared solely for the examination through focused study sessions. \\ 
         GPT Emotional & I poured my heart into studying, purely for the examination's sake. \\
         GPT Professional & My preparation was exclusively geared towards the examination, with dedicated study efforts. \\
         GPT Casual & I've only been studying for the exam, just hitting the books for that. \\\bottomrule
    \end{tabular}
    \caption{Examples of how human and systems corrected the sentences from the JFLEG dataset.}
    \label{tab:gec-samples}
\end{table*}

The authors of BEA-19 use a span-based evaluation metric $F_{0.5}$. In span-based correction, a system is only rewarded if a system edit exactly matches a reference edit in terms of both its token offsets and correction string. This is a harsh metric especially for a task like LEC where there could be so many possible variations of the correct answer to any given input. Additionally, because all of the metrics for GEC are dependent on comparing the correction with human corrected ground truth. This comparison makes an inherent assumption that human correction is the utmost form of correction, whereas from our findings, this turned out not to be true. On the contrary, we argue that because LLMs have been trained on far larger datasets than humans can ever peruse, therefore, LLM will have better estimate of the correction. In some of the cases, we even found out that GPT's LEC was far better than the ground truth corrections themselves. 

To measure the aspect or novelty of GPT that pre-existing metrics were not able to surface, we setup a preference learning task among 3 annotators where each document was annotated thrice. We sampled 100 records from both the JFLEG and BEA-19 dataset which were corrected by both GPT and then compared to corrections of humans. To assess the reliability of the annotations, we use \textit{Fleiss Kappa} score which came to 0.62 and the inter-annotator agreement (IAA) which was found to be 76\%. Furthermore, the annotators found the correction of GPT and human to be the same on average 11\% of the time. There were also occurrences when both the corrections seemed correct and therefore annotators were unable to decide, this happened 9\% of the time. As shown in table \ref{tab:ann-stats}, analysing the annotations revealed that annotators preferred the correction by GPT 73\% and 68\% respectively for JFLEG and BEA-19 datasets more than those done by human themselves. This shows the superior capabilities of GPT on tasks like LEC.

\begin{table}[h]
    \centering
    \begin{tabular}{ccccc} \toprule
        \textbf{Dataset} & \textbf{Ann 1} & \textbf{Ann 2} & \textbf{Ann 3} & \textbf{Mean}  \\ \midrule
         JFLEG           &     76.63      &       71.83     &      71.42    &     73.29      \\
         BEA-19          &     86.53      &       56.75     &      62.16    &     68.46      \\ \bottomrule
    \end{tabular}
    \caption{Preference scores of GPT over human correction for each of the three annotators.}
    \label{tab:ann-stats}
\end{table}

Furthermore, unlike pre-existing LEC systems, GPT goes beyond by offering the ability to further tune how one would like the correction to be made with respect to focus on fluency, formalism, or strict grammar and more by means of prompts. \citet{raheja2023coedit} shows the extent of different types of edits that can be formulated by LLMs. Examples of this can be seen in Table \ref{tab:gec-samples}.

On the one hand, GPT performed well on LEC and LSC, on the other hand, LLaMa, did not perform as well, even though being an LLM. We hypothesized that this might be happening due to 1) LLaMa being a decoder only architecture where GPT's embedding API might only be using encoder model 2) LLaMa not being used to operating on sentences that are short or even single sentences. We validated the latter hypothesis by setting up an experiment where we combined upto ten sentences before passing them to LLaMa for LEC. As expected, the LSC detection performance increased by 54\%.

The methodology behind how to measure performance of LEC systems goes as far back as the LEC tasks itself, one that has always been called out as not on par with judging true performance of the system \cite{bryant2023grammatical}. This is why most works ultimately aim to measure performance with human preferences. Similarly, in our study, we found GLEU and ERRANT do not paint the complete picture of GPT. Our annotators reported that the ground truth themselves were oftentimes incorrect, and as evident by preferences, annotators preferred GPT's correction over that of the humans. Another important aspect to note here, apart from the ground truth being unclean, is that our study compares supervised with unsupervised methods. Supervised models use 70\%-80\% of the available data as their training set which allows learning the nuisances of the dataset unlike GPT, where no matter the incorrect ground truths, all experiments were conducted in a zero-shot setting. Therefore, even though GPT's performance on BEA dataset lags behind state-of-the-art system, we posit that it is not due to lack of ability rather the state of art systems being trained to mimic human corrections, which in of itself are sub-optimal. 

Unlike BEA-19, JFLEG uses GLEU as its evaluation metric which is not only less stringent than $F_{0.5}$, but as our experiments show also aligns with human preferences better than $F_{0.5}$. To reiterate, in a LEC task, there could be many possible variations of the correct answer for any given input. The authors of JFLEG take this factor into consideration and provide four different variations of ground truth corrections which makes it quite suitable for modern-day generative LLMs.

\section{Conclusion \& Future Work}
Discerning whether or not a system is robust to noise and more importantly, understands semantically what a corrupted text means is the foundation of NLP. By building systems that are, for the most part, immune to noises occurring in real-world data, we make sure our NLP systems are not fragile and exhibit unintended behavior when deployed in the wild. In this work, we set out to show that modern-day LLMs do not care about corruptions as long as they are semantically the same. We do this by combining two tangential fields of NLP, Lexical Semantic Change detection and Language error correction. On the one hand, we used LSC techniques to show that the internal encoding of LLMs remains unchanged in response to corruptions in text; on the other hand, we show that unsupervised LLMs can perform zero-shot on par and even better in the downstream task of LEC. We also share preference dataset with the community. Our work paves the way for advanced LLM based LEC systems as we depart from the predominant inclusion of the LEC module as part of standard NLP systems. 

As a part of future work, we have several fronts where we strive to extend our work. First, we aim to expand our study to build and study LEC on longer passages and documents rather than just sentence-level corrections. Second, we also aim to include machine translation as part of standard LEC practices, and motivate the community at large to consider this an important step forward in the field of text normalization and future of LEC systems. Third, we aim to refine the perturbation methods as they could change the meaning of sentences at times. Finally, as discussed in Section \ref{sec:discuss}, we would like to improve the state of open source models like LLaMa to make further progress in unsupervised LEC as it has been novel in the corrections as compared to humans, as evident from our study.

\section{Limitations}
One drawback of using LLMs like GPT is the side-effects incurred i.e. unintended transformations being performed on the text. This gets even worse with smaller LLMs like LLaMa. As an example from BEA-19 dataset, a sentence \textit{``Around the city, you can find many places where people throw frigo, kitchen, "amianto", old things or furniture.''} contains a French and an Italian phrases \textit{frigo} and \textit{amianto} respectively. Even though we did not explicitly asked in the prompt shown in \ref{sec:prompt}, when GPT error corrected this, it automatically translated \textit{frigo} and \textit{amianto} to their corresponding English translations of \textit{fridge} and \textit{asbestos} respectively. This can be seen both as an advantage and disadvantage, for instance, GPT did an even superior task of LEC and implicit multi-lingual machine translation as a part of LEC task. We leave further investigation into this phenomena for future work. 


\bibliography{anthology,acl_latex}

\begin{thebibliography}{54}
\expandafter\ifx\csname natexlab\endcsname\relax\def\natexlab#1{#1}\fi

\bibitem[{Al~Sharou et~al.(2021)Al~Sharou, Li, and Specia}]{al2021towards}
Khetam Al~Sharou, Zhenhao Li, and Lucia Specia. 2021.
\newblock Towards a better understanding of noise in natural language processing.
\newblock In \emph{Proceedings of the International Conference on Recent Advances in Natural Language Processing (RANLP 2021)}, pages 53--62.

\bibitem[{Bout et~al.(2023)Bout, Podolskiy, Nikolenko, and Piontkovskaya}]{bout2023efficient}
Andrey Bout, Alexander Podolskiy, Sergey Nikolenko, and Irina Piontkovskaya. 2023.
\newblock \href {http://arxiv.org/abs/2311.11813} {Efficient grammatical error correction via multi-task training and optimized training schedule}.

\bibitem[{Brown et~al.(2020)Brown, Mann, Ryder, Subbiah, Kaplan, Dhariwal, Neelakantan, Shyam, Sastry, Askell et~al.}]{brown2020language}
Tom Brown, Benjamin Mann, Nick Ryder, Melanie Subbiah, Jared~D Kaplan, Prafulla Dhariwal, Arvind Neelakantan, Pranav Shyam, Girish Sastry, Amanda Askell, et~al. 2020.
\newblock Language models are few-shot learners.
\newblock \emph{Advances in neural information processing systems}, 33:1877--1901.

\bibitem[{Bryant et~al.(2019)Bryant, Felice, Andersen, and Briscoe}]{bryant-etal-2019-bea}
Christopher Bryant, Mariano Felice, {\O}istein~E. Andersen, and Ted Briscoe. 2019.
\newblock \href {https://doi.org/10.18653/v1/W19-4406} {The {BEA}-2019 shared task on grammatical error correction}.
\newblock In \emph{Proceedings of the Fourteenth Workshop on Innovative Use of NLP for Building Educational Applications}, pages 52--75, Florence, Italy. Association for Computational Linguistics.

\bibitem[{Bryant et~al.(2023)Bryant, Yuan, Qorib, Cao, Ng, and Briscoe}]{bryant2023grammatical}
Christopher Bryant, Zheng Yuan, Muhammad~Reza Qorib, Hannan Cao, Hwee~Tou Ng, and Ted Briscoe. 2023.
\newblock \href {http://arxiv.org/abs/2211.05166} {Grammatical error correction: A survey of the state of the art}.

\bibitem[{Devlin et~al.(2018)Devlin, Chang, Lee, and Toutanova}]{devlin2018bert}
Jacob Devlin, Ming-Wei Chang, Kenton Lee, and Kristina Toutanova. 2018.
\newblock Bert: Pre-training of deep bidirectional transformers for language understanding.
\newblock \emph{arXiv preprint arXiv:1810.04805}.

\bibitem[{Dey and Haque(2009)}]{dey2009studying}
Lipika Dey and SK~Mirajul Haque. 2009.
\newblock Studying the effects of noisy text on text mining applications.
\newblock In \emph{Proceedings of The Third Workshop on Analytics for Noisy Unstructured Text Data}, pages 107--114.

\bibitem[{Eger and Mehler(2017)}]{eger2017linearity}
Steffen Eger and Alexander Mehler. 2017.
\newblock On the linearity of semantic change: Investigating meaning variation via dynamic graph models.
\newblock \emph{arXiv preprint arXiv:1704.02497}.

\bibitem[{Fang et~al.(2023)Fang, Yang, Lan, Wong, Hu, Chao, and Zhang}]{fang2023chatgpt}
Tao Fang, Shu Yang, Kaixin Lan, Derek~F. Wong, Jinpeng Hu, Lidia~S. Chao, and Yue Zhang. 2023.
\newblock \href {http://arxiv.org/abs/2304.01746} {Is {ChatGPT} a highly fluent grammatical error correction system? a comprehensive evaluation}.

\bibitem[{Galliers and Sp{\"a}rck~Jones(1993)}]{galliers1993evaluating}
Julia~Rose Galliers and K~Sp{\"a}rck~Jones. 1993.
\newblock Evaluating natural language processing systems.
\newblock Technical report, University of Cambridge, Computer Laboratory.

\bibitem[{Garg et~al.(2021)Garg, Ramakrishnan, and Thumbe}]{garg2021towards}
Siddhant Garg, Goutham Ramakrishnan, and Varun Thumbe. 2021.
\newblock Towards robustness to label noise in text classification via noise modeling.
\newblock In \emph{Proceedings Of The 30th ACM International Conference On Information \& Knowledge Management}, pages 3024--3028.

\bibitem[{Gulordava and Baroni(2011)}]{gulordava2011distributional}
Kristina Gulordava and Marco Baroni. 2011.
\newblock A distributional similarity approach to the detection of semantic change in the google books ngram corpus.
\newblock In \emph{Proceedings of the GEMS 2011 workshop on geometrical models of natural language semantics}, pages 67--71.

\bibitem[{Hamilton et~al.(2016)Hamilton, Leskovec, and Jurafsky}]{hamilton2016cultural}
William~L Hamilton, Jure Leskovec, and Dan Jurafsky. 2016.
\newblock Cultural shift or linguistic drift? comparing two computational measures of semantic change.
\newblock In \emph{Proceedings of the conference on empirical methods in natural language processing. Conference on empirical methods in natural language processing}, volume 2016, page 2116. NIH Public Access.

\bibitem[{Heilman et~al.(2014)Heilman, Cahill, Madnani, Lopez, Mulholland, and Tetreault}]{heilman-etal-2014-predicting}
Michael Heilman, Aoife Cahill, Nitin Madnani, Melissa Lopez, Matthew Mulholland, and Joel Tetreault. 2014.
\newblock \href {https://doi.org/10.3115/v1/P14-2029} {Predicting grammaticality on an ordinal scale}.
\newblock In \emph{Proceedings of the 52nd Annual Meeting of the Association for Computational Linguistics (Volume 2: Short Papers)}, pages 174--180, Baltimore, Maryland. Association for Computational Linguistics.

\bibitem[{Hellrich and Hahn(2016)}]{hellrich2016bad}
Johannes Hellrich and Udo Hahn. 2016.
\newblock Bad company—neighborhoods in neural embedding spaces considered harmful.
\newblock In \emph{Proceedings of coling 2016, the 26th international conference on computational linguistics: Technical papers}, pages 2785--2796.

\bibitem[{Kim et~al.(2014)Kim, Chiu, Hanaki, Hegde, and Petrov}]{kim2014temporal}
Yoon Kim, Yi-I Chiu, Kentaro Hanaki, Darshan Hegde, and Slav Petrov. 2014.
\newblock Temporal analysis of language through neural language models.
\newblock \emph{arXiv preprint arXiv:1405.3515}.

\bibitem[{Kisselew et~al.(2016)Kisselew, Rimell, Palmer, and Pad{\'o}}]{kisselew2016predicting}
Max Kisselew, Laura Rimell, Alexis Palmer, and Sebastian Pad{\'o}. 2016.
\newblock Predicting the direction of derivation in english conversion.
\newblock In \emph{Proceedings of the 14th sigmorphon workshop on computational research in phonetics, phonology, and morphology}, pages 93--98.

\bibitem[{Ko et~al.(2023)Ko, Chen, Das, Chuang, and Daniel}]{ko2023robustness}
Ching-Yun Ko, Pin-Yu Chen, Payel Das, Yung-Sung Chuang, and Luca Daniel. 2023.
\newblock On robustness-accuracy characterization of large language models using synthetic datasets.
\newblock In \emph{International Conference on Machine Learning}.

\bibitem[{Kojima et~al.(2022)Kojima, Gu, Reid, Matsuo, and Iwasawa}]{kojima2022large}
Takeshi Kojima, Shixiang~Shane Gu, Machel Reid, Yutaka Matsuo, and Yusuke Iwasawa. 2022.
\newblock Large language models are zero-shot reasoners.
\newblock \emph{Advances in neural information processing systems}, 35:22199--22213.

\bibitem[{Lopresti(2008)}]{lopresti2008optical}
Daniel Lopresti. 2008.
\newblock Optical character recognition errors and their effects on natural language processing.
\newblock In \emph{Proceedings of the second workshop on Analytics for Noisy Unstructured Text Data}, pages 9--16.

\bibitem[{Ma(2019)}]{ma2019nlpaug}
Edward Ma. 2019.
\newblock Nlp augmentation.
\newblock https://github.com/makcedward/nlpaug.

\bibitem[{Maas et~al.(2011)Maas, Daly, Pham, Huang, Ng, and Potts}]{maas-etal-2011-learning}
Andrew~L. Maas, Raymond~E. Daly, Peter~T. Pham, Dan Huang, Andrew~Y. Ng, and Christopher Potts. 2011.
\newblock \href {https://aclanthology.org/P11-1015} {Learning word vectors for sentiment analysis}.
\newblock In \emph{Proceedings of the 49th Annual Meeting of the Association for Computational Linguistics: Human Language Technologies}, pages 142--150, Portland, Oregon, USA. Association for Computational Linguistics.

\bibitem[{Malykh(2019)}]{malykh-2019-robust}
Valentin Malykh. 2019.
\newblock \href {https://doi.org/10.18653/v1/P19-2002} {Robust to noise models in natural language processing tasks}.
\newblock In \emph{Proceedings of the 57th Annual Meeting of the Association for Computational Linguistics: Student Research Workshop}, pages 10--16, Florence, Italy. Association for Computational Linguistics.

\bibitem[{Martinc et~al.(2019)Martinc, Novak, and Pollak}]{martinc2019leveraging}
Matej Martinc, Petra~Kralj Novak, and Senja Pollak. 2019.
\newblock Leveraging contextual embeddings for detecting diachronic semantic shift.
\newblock \emph{arXiv preprint arXiv:1912.01072}.

\bibitem[{Mikolov et~al.(2013)Mikolov, Sutskever, Chen, Corrado, and Dean}]{mikolov2013distributed}
Tomas Mikolov, Ilya Sutskever, Kai Chen, Greg~S Corrado, and Jeff Dean. 2013.
\newblock Distributed representations of words and phrases and their compositionality.
\newblock \emph{Advances in neural information processing systems}, 26.

\bibitem[{Montariol et~al.(2021)Montariol, Martinc, and Pivovarova}]{montariol-etal-2021-scalable}
Syrielle Montariol, Matej Martinc, and Lidia Pivovarova. 2021.
\newblock \href {https://doi.org/10.18653/v1/2021.naacl-main.369} {Scalable and interpretable semantic change detection}.
\newblock In \emph{Proceedings of the 2021 Conference of the North American Chapter of the Association for Computational Linguistics: Human Language Technologies}, pages 4642--4652, Online. Association for Computational Linguistics.

\bibitem[{N{\'a}plava et~al.(2021)N{\'a}plava, Popel, Straka, and Strakov{\'a}}]{naplava2021understanding}
Jakub N{\'a}plava, Martin Popel, Milan Straka, and Jana Strakov{\'a}. 2021.
\newblock Understanding model robustness to user-generated noisy texts.
\newblock \emph{arXiv preprint arXiv:2110.07428}.

\bibitem[{Napoles et~al.(2017)Napoles, Sakaguchi, and Tetreault}]{napoles-etal-2017-jfleg}
Courtney Napoles, Keisuke Sakaguchi, and Joel Tetreault. 2017.
\newblock \href {https://aclanthology.org/E17-2037} {{JFLEG}: A fluency corpus and benchmark for grammatical error correction}.
\newblock In \emph{Proceedings of the 15th Conference of the {E}uropean Chapter of the Association for Computational Linguistics: Volume 2, Short Papers}, pages 229--234, Valencia, Spain. Association for Computational Linguistics.

\bibitem[{Ng et~al.(2014)Ng, Wu, Briscoe, Hadiwinoto, Susanto, and Bryant}]{ng-etal-2014-conll}
Hwee~Tou Ng, Siew~Mei Wu, Ted Briscoe, Christian Hadiwinoto, Raymond~Hendy Susanto, and Christopher Bryant. 2014.
\newblock \href {https://doi.org/10.3115/v1/W14-1701} {The {C}o{NLL}-2014 shared task on grammatical error correction}.
\newblock In \emph{Proceedings of the Eighteenth Conference on Computational Natural Language Learning: Shared Task}, pages 1--14, Baltimore, Maryland. Association for Computational Linguistics.

\bibitem[{OpenAI(2023)}]{openai2023gpt4}
OpenAI. 2023.
\newblock \href {http://arxiv.org/abs/2303.08774} {G{PT}-4 technical report}.

\bibitem[{Passonneau et~al.(2009)Passonneau, Rudin, Radeva, and Liu}]{passonneau2009reducing}
Rebecca~J Passonneau, Cynthia Rudin, Axinia Radeva, and Zhi~An Liu. 2009.
\newblock Reducing noise in labels and features for a real world dataset: Application of nlp corpus annotation methods.
\newblock In \emph{Computational Linguistics and Intelligent Text Processing: 10th International Conference, CICLing 2009, Mexico City, Mexico, March 1-7, 2009. Proceedings 10}, pages 86--97. Springer.

\bibitem[{Qiu and Xu(2022)}]{qiu2022histbert}
Wenjun Qiu and Yang Xu. 2022.
\newblock Histbert: A pre-trained language model for diachronic lexical semantic analysis.
\newblock \emph{arXiv preprint arXiv:2202.03612}.

\bibitem[{Raheja et~al.(2023)Raheja, Kumar, Koo, and Kang}]{raheja2023coedit}
Vipul Raheja, Dhruv Kumar, Ryan Koo, and Dongyeop Kang. 2023.
\newblock Coedit: Text editing by task-specific instruction tuning.
\newblock \emph{arXiv preprint arXiv:2305.09857}.

\bibitem[{Rosenfeld and Erk(2018)}]{rosenfeld2018deep}
Alex Rosenfeld and Katrin Erk. 2018.
\newblock Deep neural models of semantic shift.
\newblock In \emph{Proceedings of the 2018 Conference of the North American Chapter of the Association for Computational Linguistics: Human Language Technologies, Volume 1 (Long Papers)}, pages 474--484.

\bibitem[{Rothe et~al.(2021)Rothe, Mallinson, Malmi, Krause, and Severyn}]{rothe2021simple}
Sascha Rothe, Jonathan Mallinson, Eric Malmi, Sebastian Krause, and Aliaksei Severyn. 2021.
\newblock A simple recipe for multilingual grammatical error correction.
\newblock \emph{arXiv preprint arXiv:2106.03830}.

\bibitem[{Schlechtweg et~al.(2019)Schlechtweg, H{\"a}tty, Del~Tredici, and Walde}]{schlechtweg2019wind}
Dominik Schlechtweg, Anna H{\"a}tty, Marco Del~Tredici, and Sabine Schulte~im Walde. 2019.
\newblock A wind of change: Detecting and evaluating lexical semantic change across times and domains.
\newblock \emph{arXiv preprint arXiv:1906.02979}.

\bibitem[{Schlechtweg et~al.(2020)Schlechtweg, McGillivray, Hengchen, Dubossarsky, and Tahmasebi}]{schlechtweg2020semeval}
Dominik Schlechtweg, Barbara McGillivray, Simon Hengchen, Haim Dubossarsky, and Nina Tahmasebi. 2020.
\newblock Semeval-2020 task 1: Unsupervised lexical semantic change detection.
\newblock \emph{arXiv preprint arXiv:2007.11464}.

\bibitem[{Shoemark et~al.(2020)Shoemark, Liza, Nguyen, Hale, and McGillivray}]{shoemark2020room}
Philippa Shoemark, Farhana~Ferdousi Liza, Dong Nguyen, Scott Hale, and Barbara McGillivray. 2020.
\newblock Room to glo: A systematic comparison of semantic change detection approaches with word embeddings.
\newblock In \emph{Proceedings of the 2019 Conference on Empirical Methods in Natural Language Processing and the 9th International Joint Conference on Natural Language Processing (EMNLP-IJCNLP)}, pages 66--76. Association for Computational Linguistics.

\bibitem[{Sidorov et~al.(2013)Sidorov, Gupta, Tozer, Catala, Catena, and Fuentes}]{sidorov2013rule}
Grigori Sidorov, Anubhav Gupta, Martin Tozer, Dolors Catala, Angels Catena, and Sandrine Fuentes. 2013.
\newblock Rule-based system for automatic grammar correction using syntactic n-grams for english language learning (l2).
\newblock In \emph{Proceedings of the Seventeenth Conference on Computational Natural Language Learning: Shared Task}, pages 96--101.

\bibitem[{Srivastava et~al.(2020)Srivastava, Makhija, and Gupta}]{srivastava2020noisy}
Ankit Srivastava, Piyush Makhija, and Anuj Gupta. 2020.
\newblock Noisy text data: Achilles’ heel of bert.
\newblock In \emph{Proceedings of the Sixth Workshop on Noisy User-generated Text (W-NUT 2020)}, pages 16--21.

\bibitem[{Stahlberg and Kumar(2021)}]{stahlberg2021synthetic}
Felix Stahlberg and Shankar Kumar. 2021.
\newblock Synthetic data generation for grammatical error correction with tagged corruption models.
\newblock \emph{arXiv preprint arXiv:2105.13318}.

\bibitem[{Tahmasebi et~al.(2021)Tahmasebi, Borin, and Jatowt}]{tahmasebi2021survey}
Nina Tahmasebi, Lars Borin, and Adam Jatowt. 2021.
\newblock Survey of computational approaches to lexical semantic change detection.
\newblock \emph{Computational approaches to semantic change}, 6(1).

\bibitem[{Tarnavskyi et~al.(2022)Tarnavskyi, Chernodub, and Omelianchuk}]{tarnavskyi2022ensembling}
Maksym Tarnavskyi, Artem Chernodub, and Kostiantyn Omelianchuk. 2022.
\newblock Ensembling and knowledge distilling of large sequence taggers for grammatical error correction.
\newblock \emph{arXiv preprint arXiv:2203.13064}.

\bibitem[{Touvron et~al.(2023)Touvron, Lavril, Izacard, Martinet, Lachaux, Lacroix, Rozi{\`e}re, Goyal, Hambro, Azhar et~al.}]{touvron2023llama}
Hugo Touvron, Thibaut Lavril, Gautier Izacard, Xavier Martinet, Marie-Anne Lachaux, Timoth{\'e}e Lacroix, Baptiste Rozi{\`e}re, Naman Goyal, Eric Hambro, Faisal Azhar, et~al. 2023.
\newblock Llama: Open and efficient foundation language models.
\newblock \emph{arXiv preprint arXiv:2302.13971}.

\bibitem[{Vaswani et~al.(2017)Vaswani, Shazeer, Parmar, Uszkoreit, Jones, Gomez, Kaiser, and Polosukhin}]{vaswani2017attention}
Ashish Vaswani, Noam Shazeer, Niki Parmar, Jakob Uszkoreit, Llion Jones, Aidan~N Gomez, {\L}ukasz Kaiser, and Illia Polosukhin. 2017.
\newblock Attention is all you need.
\newblock \emph{Advances in neural information processing systems}, 30.

\bibitem[{Wang et~al.(2023)Wang, Ma, Yu, Gui, Zhang, Huang, Ma, Chang, Zhang, Shen, Wang, Zhao, and Tao}]{wang2023large}
Haoyu Wang, Guozheng Ma, Cong Yu, Ning Gui, Linrui Zhang, Zhiqi Huang, Suwei Ma, Yongzhe Chang, Sen Zhang, Li~Shen, Xueqian Wang, Peilin Zhao, and Dacheng Tao. 2023.
\newblock \href {http://arxiv.org/abs/2309.11166} {Are large language models really robust to word-level perturbations?}

\bibitem[{Wei et~al.(2022)Wei, Wang, Schuurmans, Bosma, Xia, Chi, Le, Zhou et~al.}]{wei2022chain}
Jason Wei, Xuezhi Wang, Dale Schuurmans, Maarten Bosma, Fei Xia, Ed~Chi, Quoc~V Le, Denny Zhou, et~al. 2022.
\newblock Chain-of-thought prompting elicits reasoning in large language models.
\newblock \emph{Advances in Neural Information Processing Systems}, 35:24824--24837.

\bibitem[{Wu et~al.(2021)Wu, Chen, Ding, and Tao}]{wu2021bridging}
Di~Wu, Yiren Chen, Liang Ding, and Dacheng Tao. 2021.
\newblock Bridging the gap between clean data training and real-world inference for spoken language understanding.
\newblock \emph{arXiv preprint arXiv:2104.06393}.

\bibitem[{Xing et~al.(2013)Xing, Wang, Wong, Chao, and Zeng}]{xing2013checker}
Junwen Xing, Longyue Wang, Derek~F Wong, Lidia~S Chao, and Xiaodong Zeng. 2013.
\newblock Um-checker: A hybrid system for english grammatical error correction.
\newblock In \emph{Proceedings of the Seventeenth Conference on Computational Natural Language Learning: Shared Task}, pages 34--42.

\bibitem[{Xu and Kemp(2015)}]{xu2015computational}
Yang Xu and Charles Kemp. 2015.
\newblock A computational evaluation of two laws of semantic change.
\newblock In \emph{CogSci}.

\bibitem[{Yannakoudakis et~al.(2011)Yannakoudakis, Briscoe, and Medlock}]{yannakoudakis-etal-2011-new}
Helen Yannakoudakis, Ted Briscoe, and Ben Medlock. 2011.
\newblock \href {https://aclanthology.org/P11-1019} {A new dataset and method for automatically grading {ESOL} texts}.
\newblock In \emph{Proceedings of the 49th Annual Meeting of the Association for Computational Linguistics: Human Language Technologies}, pages 180--189, Portland, Oregon, USA. Association for Computational Linguistics.

\bibitem[{Zhang et~al.(2022)Zhang, Zhang, Li, Bao, Li, and Zhang}]{zhang2022syngec}
Yue Zhang, Bo~Zhang, Zhenghua Li, Zuyi Bao, Chen Li, and Min Zhang. 2022.
\newblock Syngec: Syntax-enhanced grammatical error correction with a tailored gec-oriented parser.
\newblock \emph{arXiv preprint arXiv:2210.12484}.

\bibitem[{Zhou et~al.(2022{\natexlab{a}})Zhou, Sch{\"a}rli, Hou, Wei, Scales, Wang, Schuurmans, Cui, Bousquet, Le et~al.}]{zhou2022least}
Denny Zhou, Nathanael Sch{\"a}rli, Le~Hou, Jason Wei, Nathan Scales, Xuezhi Wang, Dale Schuurmans, Claire Cui, Olivier Bousquet, Quoc Le, et~al. 2022{\natexlab{a}}.
\newblock Least-to-most prompting enables complex reasoning in large language models.
\newblock \emph{arXiv preprint arXiv:2205.10625}.

\bibitem[{Zhou et~al.(2022{\natexlab{b}})Zhou, Muresanu, Han, Paster, Pitis, Chan, and Ba}]{zhou2022large}
Yongchao Zhou, Andrei~Ioan Muresanu, Ziwen Han, Keiran Paster, Silviu Pitis, Harris Chan, and Jimmy Ba. 2022{\natexlab{b}}.
\newblock Large language models are human-level prompt engineers.
\newblock \emph{arXiv preprint arXiv:2211.01910}.

\end{thebibliography}




         

\end{document}